# IMAGE COMPRESSION BY EMBEDDING FIVE MODULUS METHOD INTO JPEG

Firas A. Jassim

Management Information Systems Department,
Irbid National University, Irbid 2600, Jordan
Firasajil@yahoo.com

***ABSTRACT***

*The standard JPEG format is almost the optimum format in image compression. The compression ratio in JPEG sometimes reaches 30:1. The compression ratio of JPEG could be increased by embedding the Five Modulus Method (FMM) into the JPEG algorithm. The novel algorithm gives twice the time as the standard JPEG algorithm or more. The novel algorithm was called FJPEG (Five-JPEG). The quality of the reconstructed image after compression is approximately approaches the JPEG. Standard test images have been used to support and implement the suggested idea in this paper and the error metrics have been computed and compared with JPEG.*

## 1. INTRODUCTION

The main goal of image compression methods is to represent the original images with fewer bits. Recently, image compression is very popular in many research areas. According to the research area, one of the two types of compression, which are lossless and lossy, can be used [13]. Lossless compression can retrieve the original image after reconstruction. Since it is impossible to compress the image with high compression ratio without errors, therefore; lossy image compression was used to obtain high compression ratios. Consequently, reducing image size with lossy image compression gives much more convenient ratio than lossless image compression [7] [9].

Since the mid of the eighties of the last century, the International Telecommunication Union (ITU) and the International Organization for Standardization (ISO) have been working together to obtain a standard compression image extension for still images. The recommendation ISO DIS 10918-1 known as JPEG Joint Photographic Experts Group [11]. Digital Compression and Coding of Continuous-tone Still Images and also ITU-T Recommendation T.81 [11]. After comparing many coding schemes for image compression, the JPEG members selected a Discrete Cosine Transform (DCT). JPEG became a Draft International Standard (DIS) in 1991 and an International Standard (IS) in 1992 [12]. JPEG has become an international standard for lossy compression of digital image.

## 2. FIVE MODULUS METHOD

Five modulus method (FMM) was first introduced by [6]. The main concept of this method is to convert the value of each pixel into multiples of five. This conversion omits parts of the signal that will not be noticed by the signal receiver namely the Human Visual System (HVS). Since the neighbouring pixels are correlated in image matrix, therefore; finding less correlated representation of image is one of the most important tasks. The main principle of image compression states that the neighbours of a pixel tend to have the same immediate neighbours [4].

Hence, the FMM technique tends to divide image into 8×8 blocks. After that, each pixel in every block can be transformed into a number divisible by 5. The effectiveness of this transformation will not be noticed by the Human Visual System (HVS) [6]. Therefore, each pixel value is from the multiples of 5 only, i.e. 0, 5, 10, 15, 20, … , 255. The FMM algorithm could be stated as:

*if Pixel value Mod 5 = 4*
    *Pixel value=Pixel value+1*
*if Pixel value Mod 5 = 3*
    *Pixel value=Pixel value+2*
*if Pixel value Mod 5 = 2*
    *Pixel value=Pixel value-2*
*if Pixel value Mod 5 = 1*
    *Pixel value=Pixel value-1*

Now, to illustrate the method of Five Modulus Method (FMM). An arbitrary 8×8 block has been taken randomly from an arbitrary digital image and showed in table 1.

Table 1. Original 8×8 block

| | | | | | | | |
|---|---|---|---|---|---|---|---|
| 106 | 98 | 104 | 102 | 109 | 110 | 107 | 113 |
| 103 | 107 | 104 | 110 | 109 | 110 | 110 | 113 |
| 106 | 106 | 105 | 110 | 111 | 107 | 104 | 108 |
| 104 | 105 | 110 | 111 | 109 | 108 | 110 | 104 |
| 106 | 106 | 119 | 113 | 111 | 107 | 109 | 108 |
| 106 | 104 | 101 | 105 | 104 | 104 | 107 | 113 |
| 97 | 103 | 104 | 101 | 102 | 104 | 106 | 110 |
| 103 | 106 | 110 | 105 | 103 | 105 | 103 | 108 |

After that, the FMM algorithm shown earlier may be applied to the 8×8 block in table (1). Every pixel value was converted into multiple of five, i.e. the first pixel which is (106) may be converted into (105), etc. Therefore, the new resulting 8×8 block was showed in table (2).

Table 2. Converting 8×8 block by five modulus method (FMM)

| | | | | | | | |
|---|---|---|---|---|---|---|---|
| 105 | 100 | 105 | 100 | 110 | 110 | 105 | 115 |
| 105 | 105 | 105 | 110 | 110 | 110 | 110 | 115 |
| 105 | 105 | 105 | 110 | 110 | 105 | 105 | 110 |
| 105 | 105 | 110 | 110 | 110 | 110 | 110 | 105 |
| 105 | 105 | 120 | 115 | 110 | 105 | 110 | 110 |
| 105 | 105 | 100 | 105 | 105 | 105 | 105 | 115 |
| 95 | 105 | 105 | 100 | 100 | 105 | 105 | 110 |
| 105 | 105 | 110 | 105 | 105 | 105 | 105 | 110 |

Now, to complete the FFM method, the 8×8 block shown in table (2) may be divided by 5 to reduce the pixel values into a lesser values. Therefore, the first converted pixel (105) would be (105/5=21), etc. The evaluated 8×8 block after division is shown in table (3).

Table 3. Dividing 8×8 block in table (1) by 5

| | | | | | | | |
|---|---|---|---|---|---|---|---|
| 21 | 20 | 21 | 20 | 22 | 22 | 21 | 23 |
| 21 | 21 | 21 | 22 | 22 | 22 | 22 | 23 |
| 21 | 21 | 21 | 22 | 22 | 21 | 21 | 22 |
| 21 | 21 | 22 | 22 | 22 | 22 | 22 | 21 |
| 21 | 21 | 24 | 23 | 22 | 21 | 22 | 22 |
| 21 | 21 | 20 | 21 | 21 | 21 | 21 | 23 |
| 19 | 21 | 21 | 20 | 20 | 21 | 21 | 22 |
| 21 | 21 | 22 | 21 | 21 | 21 | 21 | 22 |

The main concept of the FMM method is to reduce the dispersion (variation) between pixel values in the same 8×8 block. Hence, the standard deviation in the original 8×8 block was (3.84) while it was (0.85) in the transformed 8×8 block. This implies that, the storage space for the transformed 8×8 block will be less than that of the original 8×8 block.

## 3. FJPEG ENCODING AND DECODING

The basic technique in FJPEG encoding is to apply FMM first. Each 8×8 block is transformed into multiples of 5, i.e. Five Modulus Method (FMM), see table (2). After that, dividing the whole 8×8 block by 5 to obtain new pixel values range [0..51] which are the results of dividing [0..255] by 5. Now, the 8×8 block is ready to implement the standard JPEG exactly starting with DCT and so on. The new file format if FJPEG which is the same as JPEG but all its values are multiples of 5. Actually, the FJPEG image format could be sent over the internet or stored in the storage media, or what ever else, with a lesser file size as a compressed file. On the other hand, at the decoding side, when reconstructing the image, firstly, the IFMM (Inverse Five Modulus Method) could be applied by multiplying the 8×8 block by 5 to retrieve, approximately, the original 8×8 block. After that, the same JPEG decoding may be applied as it is. Therefore, the main contribution is this article is to embed FMM method into JPEG to reduce the file size. The whole encoding and decoding procedures may be shown in figure (1).

One of the main advantages in FJPEG compared to JPEG is that the reduction in file size is noticeable. Unfortunately, one of the main disadvantages is that on the decoding side and before applying JPEG decoding the image must be multiplied by 5. Theses calculations will be evaluated on the computer processor at the decoding side. Actually, these calculations will surely offtake some time and space from the memory and CPU.

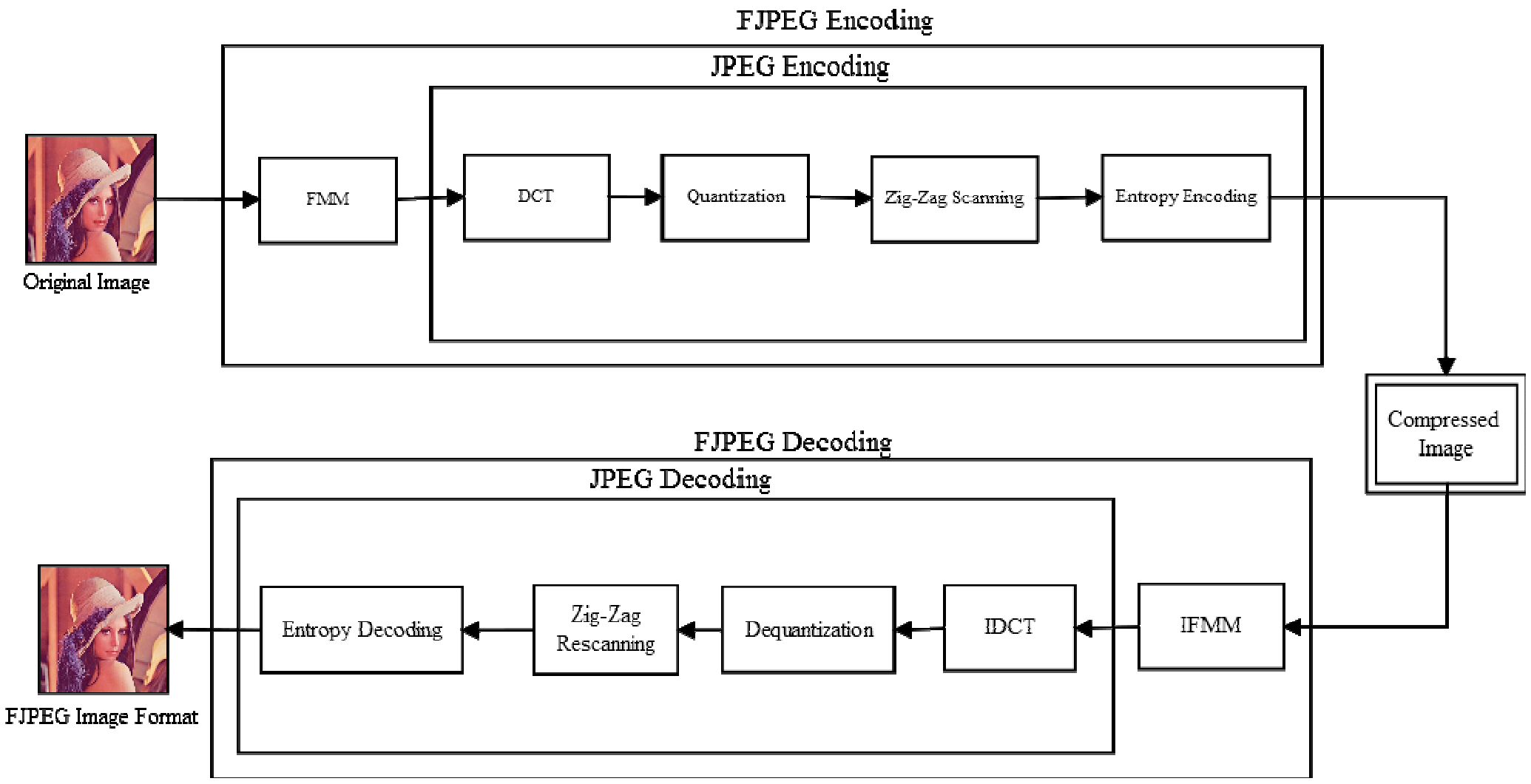

Figure 1. Encoding and Decoding procedures in FJPEG

## 4. DISCRETE COSINE TRANSFORM

Discrete Cosine Transformation (DCT), is used to compress digital images by reducing the number of pixels used to express 8x8 blocks into a lesser number of pixels. Nowadays, JPEG standard uses the DCT as its essentials. This type of lossy encoding has become the most popular transform for image compression especially in JPEG image format. The origin of the DCT back to 1974 by [1]. The DCT algorithm is completely invertible which makes it useful for both lossless and lossy image compression. The DCT transform the pixel values of the 8×8 block into two types. The first type is the highest values that can be positioned at the upper left corner of the 8×8 block. While the second type represents the values with the smaller value which can be found at the remaining areas of the 8×8 block. The coefficients of the DCT can be used to reconstruct the original 8×8 block through the Inverse Discrete Cosine Transform (IDCT) which can be used to retrieve the image from its original representation [7].

Images can be separated by DCT into segments of frequencies where less important frequencies are omitted through quantization method and the important frequencies are used to reconstruct the image through decoding technique [8]. The main two advantages for embedding FMM into JPEG are:

- Reducing the dispersion (variation) degree between DCT coefficients.
- Decreasing the number of the non-zero elements of the DCT coefficients.

According to [3], the measure DCT-STD (STD is the standard deviation) have been used to measure the clarity of am image. Here, the DCT-STD have been used to measure the difference of the DCT of the 8x8 block taken from the traditional JPEG image with 8x8 block taken from FJPEG image, tables 4 and 5 respectively.

Table 4. DCT transform of the original block

| | | | | | | | |
|---|---|---|---|---|---|---|---|
| 853 | -10 | -2 | -6 | 2 | 0 | 2 | 0 |
| 7 | -3 | -2 | 6 | 4 | 0 | 0 | 0 |
| -8 | -5 | 6 | 1 | 0 | -4 | 3 | -1 |
| 0 | -5 | 5 | 0 | 1 | 0 | 2 | 2 |
| 4 | 4 | -4 | -1 | -3 | 3 | 5 | 5 |
| -8 | -3 | 1 | 3 | 0 | 0 | 0 | 1 |
| -1 | 2 | 3 | 3 | 5 | -1 | 1 | 0 |
| 1 | 2 | -3 | -2 | 0 | -1 | 3 | 3 |

Table 5. DCT transform of the FMM block

| | | | | | | | |
|---|---|---|---|---|---|---|---|
| 170 | -2 | 0 | -1 | 0 | 0 | 0 | 0 |
| 1 | 0 | 0 | 1 | 1 | 0 | 0 | 0 |
| -1 | -1 | 1 | 0 | 0 | -1 | 0 | 0 |
| 0 | 0 | 0 | 0 | 0 | 0 | 0 | 0 |
| 1 | 0 | 0 | 0 | 0 | 0 | 0 | 1 |
| -1 | 0 | 0 | 0 | 0 | 0 | 0 | 0 |
| 0 | 0 | 0 | 0 | 1 | 0 | 0 | 0 |
| 0 | 0 | 0 | -1 | 0 | 0 | 0 | 0 |

The value of the standard deviation for table (4) is 106.65, while the standard deviation for table (5) is 21.26. Actually, when the standard deviation value is high, it means that the difference between most values and their mean values are high; and vice versa. Hence, coefficients of table (4) have high variation than table (5). It means that, the dispersion obtained by the transformed FMM is less than the traditional dispersion of the original DCT. Also, it can be seen that the number of non-zero elements of table (2) are lesser than of those of table (1). According to [5], every zero element of the DCT matrix saves operations and reduces the time complexity. Also, the total processing time required for IDCT is decreased when the number of the non-zero elements of the DCT coefficient matrix is less [10]. Actually, in image compression, few non-zero elements of the DCT coefficients could be used to represent an image [2].

Fortunately, the previous mentioned opinions decants in the behalf of the proposed idea in this article which is that the reduction of the non-zero elements of the DCT matrix will reduce the time complexity. Hence, using DCT obtained by the FMM method will gives sophisticated results than using the traditional DCT that contains more non-zero elements.

## 5. Experimental Results

Actually, the proposed FJPEG in this article has been implemented to a variety of standard test images with different spatial and frequency characteristics. As mentioned earlier, the embedding of FMM into JPEG will formulate FJPEG which has a noticeable difference in image size. Experimentally, the compression ratio between JPEG and FJPEG has been showed in table (6). Also, PSNR between JPEG and FJPEG was calculated in the same table.

Table 6. File size, CR and PSNR for test images

| | *File Size (KB)* | | *CR* | | *PSNR* | |
|---|---|---|---|---|---|---|
| | JPEG | FJPEG | JPEG | FJPEG | JPEG | FJPEG |
| Lena | 36.9 | 13.5 | 20.8:1 | 56.9:1 | 37.4882 | 32.3178 |
| Baboon | 75.6 | 27.6 | 10.2:1 | 28.8:1 | 30.7430 | 25.0817 |
| Peppers | 40.5 | 14.3 | 19.0:1 | 53.7:1 | 35.5778 | 31.6617 |
| F16 | 26 | 15 | 29.5:1 | 51.2:1 | Inf | Inf |
| Bird | 131 | 27 | 7.7:1 | 37.5:1 | 53.4068 | 30.4266 |
| ZigZag | 64.2 | 23.5 | 5.6:1 | 15.2:1 | 31.0531 | 20.1237 |
| Houses | 66.8 | 24.2 | 8.9:1 | 24.5:1 | Inf | 21.9970 |
| **Mosque** | 23.9 | 8.33 | 14.7:1 | 42.1:1 | 36.1242 | 28.6154 |

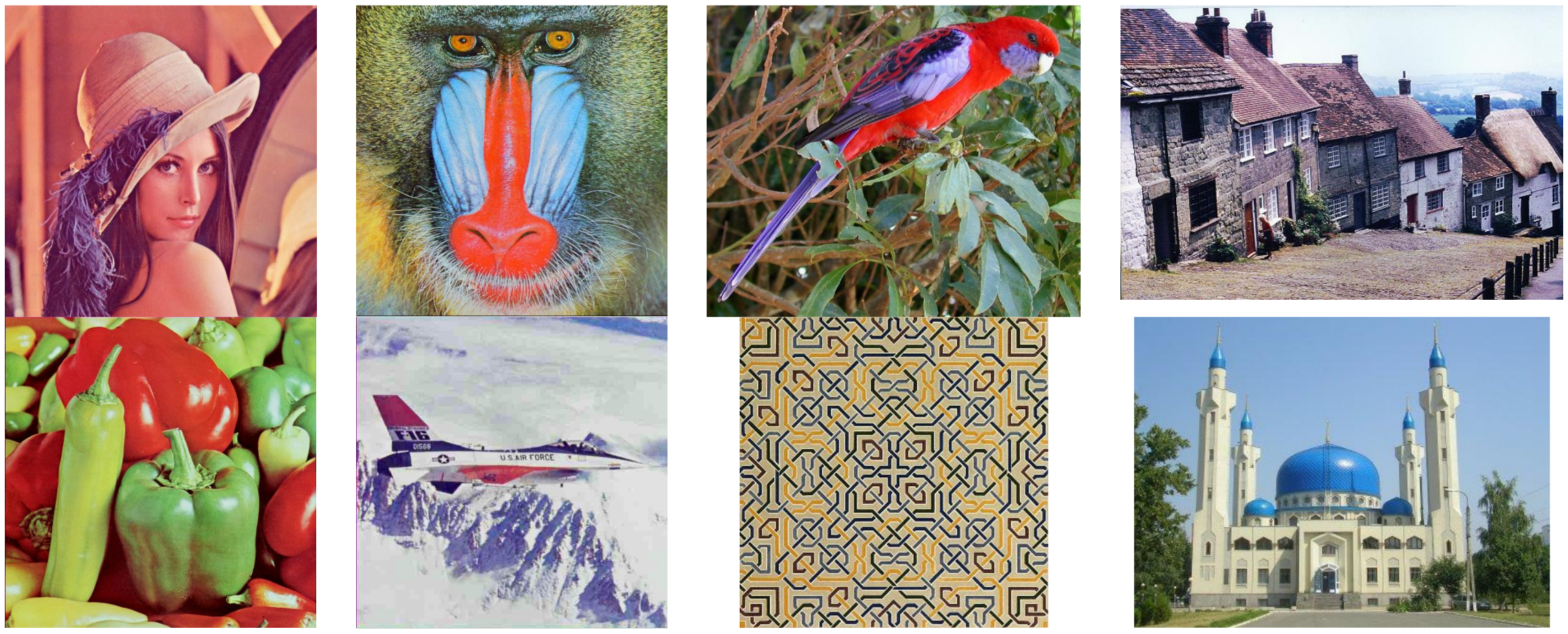

Figure 2. Variety of test images

As an example, three test images (Lena, baboon, and Peppers) have been used, figures 3,4, and 5, to show the visual and storage differences between (a) standard JPEG (b) FJPEG after encoding directly, i.e. this file will be sent over the internet and stored in the storage media which has a small file size compared to JPEG (c) FJPEG after reconstruction, i.e. multiplied by 5 on the decoding side.

As seen from figures 3,4, and 5, that the differences by the human eye are worthless and can not be recognized visually. As a quantitative measure, the PSNR has been used to compute the signal-to-noise-ratio for the original JPEG and the FJPEG and the result are presented in table (6). The differences between PSNR for the two images are reasonable. As an example, for Lena image the PSNR of the JPEG was (37.4882) while the PSNR for the FJPEG was (32.3178) which is not very high difference. Therefore, one could accept this fair difference by taking into consideration the high difference in file size (36.9 KB) for JPEG while (13.5 KB) for the FJPEG.

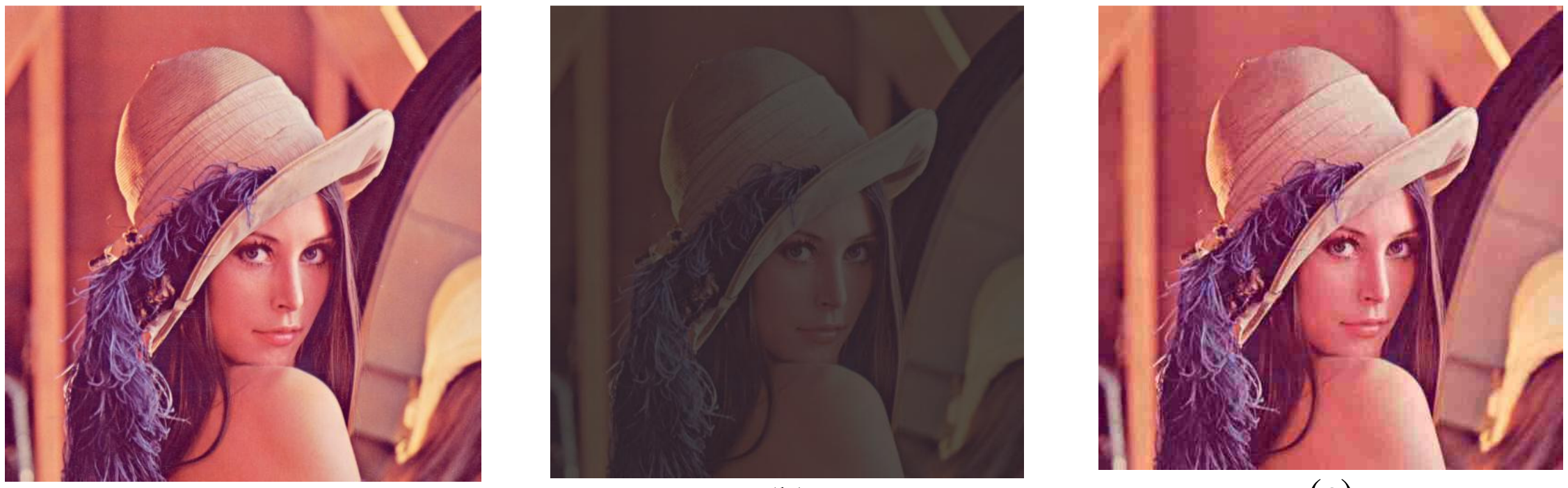

(a) (b) (c)

Figure 3. (a) JPEG (36.9 KB) (b) FJPEG (sent) (13.5 KB) (c) FJPEG×5 (reconstructed) (24 KB)

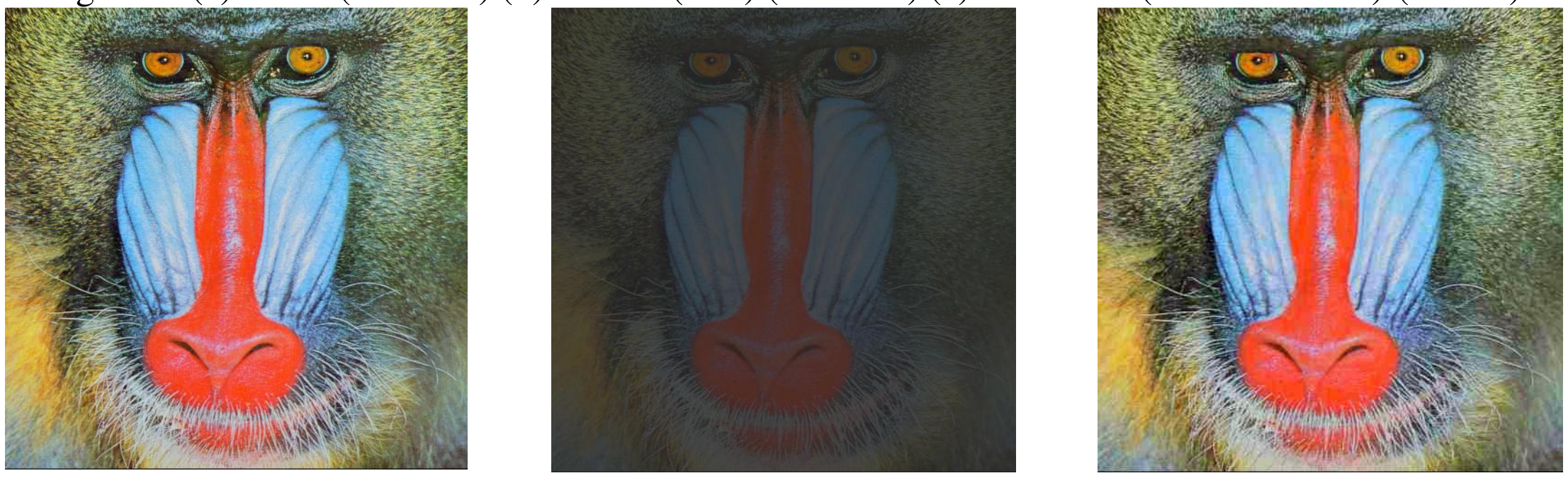

(a) (b) (c)

Figure 4. (a) JPEG (75.6 KB) (b) FJPEG (sent) (26.7 KB) (c) FJPEG×5 (reconstructed) (50.6 KB)

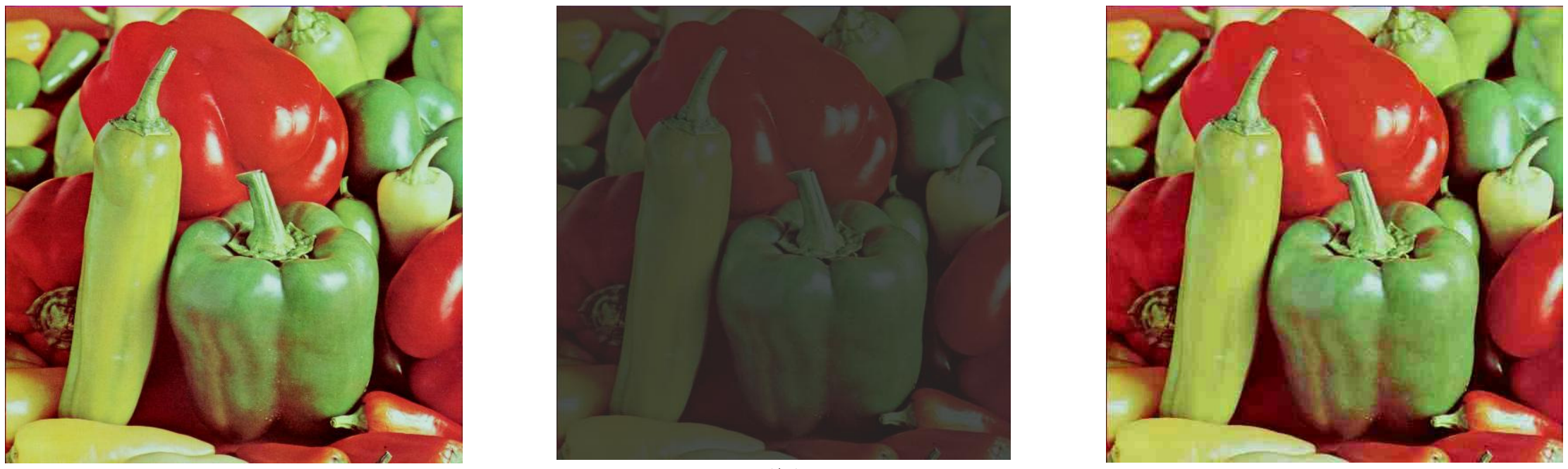

(a) (b) (c)

Figure 5. (a) JPEG (40.5 KB) (b) FJPEG (sent) (14.3 KB) (c) FJPEG×5 (reconstructed) (25.9 KB)

The main difference in compression ratio between FJPEG and JPEG may be illustrated graphically in figure (6).

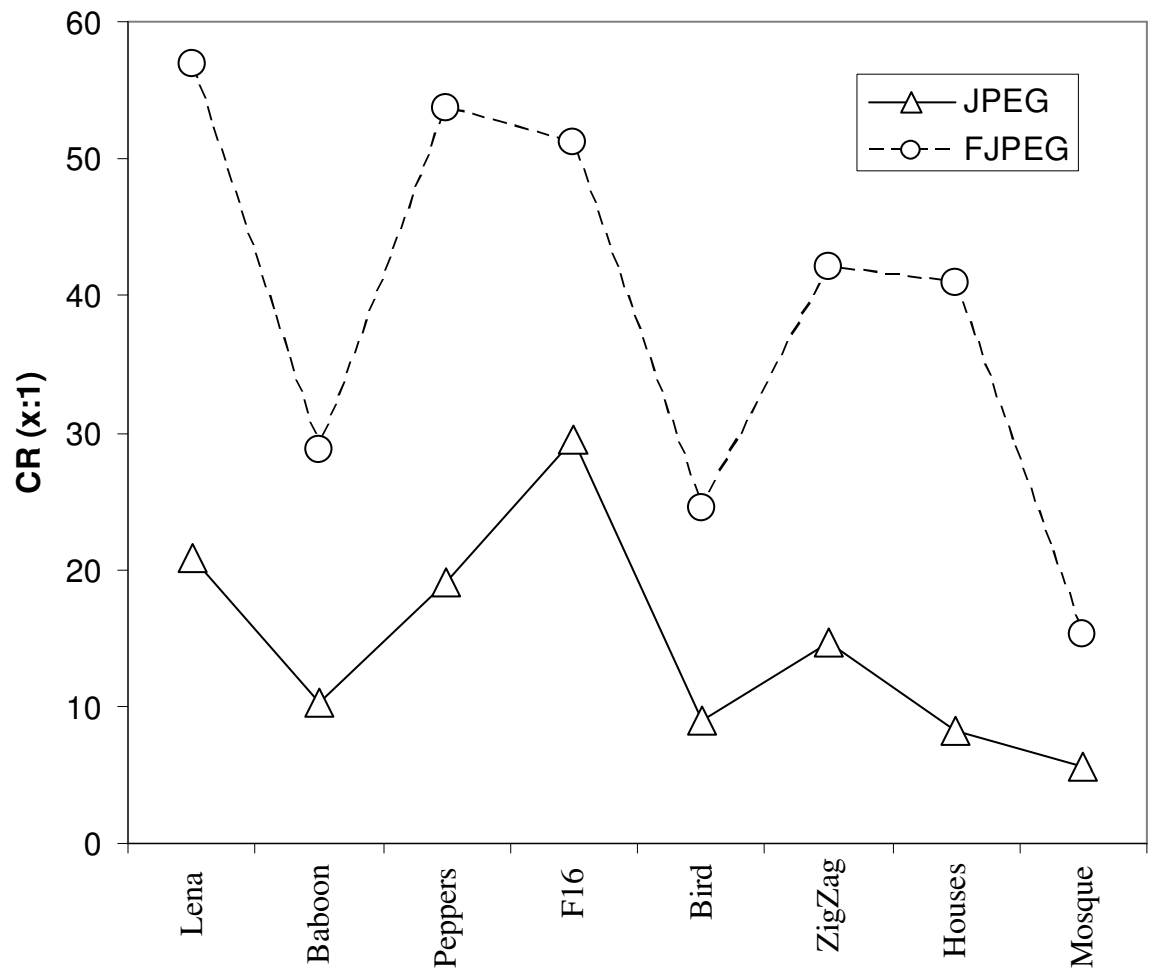

Figure 6. Difference in compression ratio between JPEG and FJPEG

## 6. CONCLUSIONS

The method of FJPEG stated in this article can be used as a redundancy alternative of the traditional JPEG in most of the disciplines of science and engineering especially in image compression because of its noticeable difference in image size or the compression ratio. As a first recommendation, the research area of the CPU time and space taken to reconstruct the FJPEG at the encoding side needs a lot of interest. As a second recommendation, the implementation of FJPEG in video compression needs to be discussed and researched. As a third recommendation, the alternative name of JPEG method in audio is called MP3, therefore; using the FJPEG in audio compression could be researched.

**Authors**

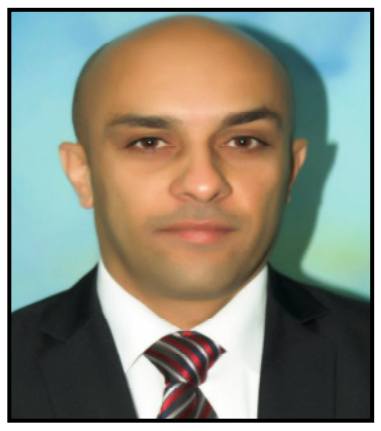

Firas Ajil Jassim was born in Baghdad, Iraq, in 1974. He received the B.S. and M.S. degrees in Applied Mathematics and Computer Applications from Al-Nahrain University, Baghdad, Iraq, in 1997 and 1999, respectively, and the Ph.D. degree in Computer Information Systems (CIS) from the Arab Academy for Banking and Financial Sciences, Amman, Jordan, in 2012. In 2012, he joined the faculty of the Department of Business Administration, college of Management Information System, Irbid National University, Irbid, Jordan, where he is currently an assistance professor. His current research interests are image compression, image interpolation, image segmentation, image enhancement, and simulation.